\documentclass[10pt]{article}
\usepackage{iclr2026_conference}
\usepackage{booktabs}
\usepackage{float}
\usepackage{graphicx}
\usepackage{amsmath}
\usepackage{multirow}
\usepackage{caption}
\usepackage{xcolor}

\title{Scaffolding the Strategist: Architecture-Dependent Reasoning Interventions in Hotelling Spatial Markets}

\author{%
    Pratyush Singh \\
    Computing + Mathematical Sciences \\
    California Institute of Technology \\
    Pasadena, CA 91125, USA \\
    \texttt{pksingh@caltech.edu}
}

\iclrfinalcopy

\begin{document}
\maketitle

\begin{abstract}
We investigate whether structured reasoning interventions improve the strategic economic reasoning of large language models, and whether their effects depend on model architecture. Using Hotelling's linear city model as a diagnostic vehicle, we evaluate GPT-4.1-mini (a standard instruction-following model) and GPT-5-mini (a reasoning-optimized model) under five conditions---an unscaffolded baseline and four reasoning interventions---across eight questions spanning deductive and abductive reasoning, three prompt framings, and three repetitions per condition, yielding 720 individually judged responses. We find a statistically significant crossover interaction between scaffolding type and model architecture ($t(7) = 4.79$, $p = 0.002$, $d = 1.69$): commitment scaffolding improves the standard model ($+0.21$) while degrading the reasoning model ($-0.63$), and principled separation shows the opposite pattern ($-0.40$ vs.\ $+0.31$). Both crossovers are individually significant (commitment: $p = 0.040$; separation: $p = 0.002$) and hold across all eight questions with 7/8 directional consistency. Adversarial stress-testing harms both models, with $2.6\times$ greater degradation for the reasoning model ($-1.47$ vs.\ $-0.57$; $p = 0.038$), and the damage correlates negatively with baseline difficulty ($R^2 = 0.36$, $p = 0.014$). We further document a persistent declarative--procedural gap in which both models identify correct strategies at rates far exceeding their ability to execute them; separation fully closes this gap for the reasoning model while no intervention helps the standard model.
\end{abstract}

% ===================================================================
\section{Introduction}
\label{sec:intro}
% ===================================================================

Large language models have achieved remarkable performance on mathematical and logical benchmarks \citep{openai2024gpt4}, yet their capacity for strategic economic reasoning---where agents must integrate domain knowledge, anticipate competitors, and translate understanding into concrete actions---remains poorly understood. Real-world decision-making requires multi-step strategic inference: reasoning backward from outcomes, maintaining consistency across choices, and distinguishing equilibrium predictions from naive intuitions.

A growing body of work has begun to address this gap by embedding LLM evaluation within game-theoretic frameworks. Studies using Cournot competition \citep{lin2024strategic}, repeated behavioral games \citep{akata2023playing}, and decomposed strategic reasoning \citep{gandhi2023strategic} have consistently found that LLMs exhibit systematic biases invisible to standard benchmarks. GTBench \citep{duan2024gtbench} provides the most comprehensive game-theoretic LLM benchmark to date, spanning ten complete- and incomplete-information games, while EconArena \citep{guo2024econarena} benchmarks economic reasoning using equilibrium deviation as a primary metric. However, these evaluations rely predominantly on discrete action spaces and simultaneous-move games, leaving continuous strategy spaces, multi-stage backward induction, and counterintuitive equilibria unexplored.

A critical dimension also remains unaddressed: how reasoning interventions interact with model architecture. The deployment of reasoning-optimized models with built-in chain-of-thought has raised questions about whether external scaffolding retains its value. \citet{sprague2024cot} found diminishing returns from chain-of-thought prompting on models that already reason internally, and GTBench similarly reported that chain-of-thought does not always help in game-theoretic settings. Meanwhile, \citet{zhang2025splitbrain} coined \emph{computational split-brain syndrome} to describe models that explain principles they cannot follow, a declarative--procedural gap that \citet{gandhi2023strategic} showed varies systematically across reasoning subtypes. If a model already conducts internal deliberation, does external scaffolding help, do nothing, or actively interfere?

We address these questions using Hotelling's linear city model \citep{hotelling1929stability} as a diagnostic vehicle. We evaluate GPT-4.1-mini (a standard instruction-following model) and GPT-5-mini (a reasoning-optimized model) under five conditions: an unscaffolded baseline and four reasoning interventions. Each condition is tested across three prompt framings and three repetitions, yielding 720 individually judged responses scored by GPT-5.2 as an automated evaluator with human-validated calibration ($\kappa = 0.97$).

Our contributions are fourfold. First, we document a statistically significant crossover interaction between scaffolding type and architecture ($p = 0.002$, $d = 1.69$). Second, we show that adversarial stress-testing universally degrades performance, with greater harm to the reasoning model ($p = 0.038$) that increases with baseline competence ($R^2 = 0.36$). Third, we quantify the declarative--procedural gap in economic reasoning, showing principled separation fully closes it for the reasoning model while no intervention helps the standard model. Fourth, we validate Hotelling spatial competition as a contamination-resistant diagnostic framework with a graduated difficulty scale.

% ===================================================================
\section{Related Work}
\label{sec:related}
% ===================================================================

The study of LLM behavior in game-theoretic contexts has grown rapidly. \citet{horton2023homo} framed LLMs as "Homo Silicus," demonstrating recognizable economic preferences, while subsequent work documented emergent strategic behaviors: supracompetitive pricing without collusion instructions \citep{fish2024algorithmic}, market division in Cournot competition \citep{lin2024strategic}, and deviations from Nash play in repeated games \citep{akata2023playing}. GTBench \citep{duan2024gtbench} systematized evaluations across ten games, finding inconsistent benefits from chain-of-thought, and EconArena \citep{guo2024econarena} extended the approach to multi-agent market simulations. Our work differs in using continuous strategy spaces, testing reasoning interventions rather than comparing models, and employing a factorial design that reveals interaction effects.

The question of how to scaffold LLM reasoning connects to the chain-of-thought literature. Since \citet{wei2022cot} established chain-of-thought prompting, \citet{sprague2024cot} have shown its benefits concentrate in mathematical and symbolic tasks, with diminishing returns elsewhere. Our interventions extend beyond simple chain-of-thought by imposing structural constraints---committing to premises or separating principle identification from application---and the crossover we document suggests these interact with built-in reasoning in ways uniform prompting cannot capture.

A parallel line of work has examined the gap between what LLMs know and what they can do. \citet{zhang2025splitbrain} identified "computational split-brain syndrome," where models articulate correct principles but fail to apply them, and \citet{gandhi2023strategic} showed this gap varies across reasoning subtypes, with abductive tasks exhibiting larger gaps than deductive ones. In human cognition, the distinction between knowing-that and knowing-how has a long history \citep{ryle1949concept}, and the verbal overshadowing effect \citep{schooler1990verbal} shows that explicit articulation can degrade performance on tasks that benefit from implicit processing. Our declarative--procedural analysis connects these threads: we show the gap has qualitatively different origins in different architectures and responds differently to intervention.

% ===================================================================
\section{Why Hotelling}
\label{sec:hotelling}
% ===================================================================

The choice of evaluation domain is deliberate. We require a domain that demands multi-step strategic reasoning with verifiable ground truth, separates reasoning subtypes within a unified framework, resists training-data contamination, and exposes the declarative--procedural gap. Hotelling's spatial competition model \citep{hotelling1929stability} satisfies all four requirements.

Unlike one-shot simultaneous games, Hotelling competition requires solving a two-stage sequential game: firms first choose locations, then compete on prices. Solving for subgame-perfect equilibrium demands backward induction---deriving price equilibrium as a function of locations, then optimizing location choice. By varying information structure, we construct both deductive tasks (deriving equilibrium outcomes from axioms) and abductive tasks (inferring hidden parameters from observed outcomes). Standard logic benchmarks test these in isolation; Hotelling unifies them under shared conceptual vocabulary and verifiable ground truth.

The model also features continuous location and price spaces rather than discrete action sets. It produces counterintuitive equilibria---for instance, maximum differentiation under quadratic transport costs contradicts the naive intuition that firms should cluster near customers---that distinguish genuine understanding from pattern matching. Our non-standard parameterizations (asymmetric costs, partial information, multi-segment markets) are unlikely to appear in any training corpus \citep{lin2024strategic}.

Finally, Hotelling naturally separates "knowing what to do" from "doing it." A model might correctly identify that a firm should differentiate maximally yet fail to compute the equilibrium location or price. Our Component~B abductive tasks expose this gap by requiring application of recognized principles to extract hidden information from observed data.

% ===================================================================
\section{Experiments}
\label{sec:experiments}
% ===================================================================

\subsection{Task Design}

We design eight questions organized into two components reflecting the distinction between deductive and abductive reasoning.

Component~A contains five deductive reasoning questions requiring forward derivation from given premises. A1 tests parameter sensitivity (equilibrium prices and profits as transport costs scale). A2 tests location change consequences (effects of halving inter-firm distance). A3 tests cost structure identification (classifying transport cost type from market share data). A4 tests asymmetric cost analysis (equilibrium predictions under a cost shock). A5 tests entry deterrence (computing entry profitability and designing a deterrence strategy via backward induction).

Component~B contains three abductive reasoning questions requiring inference from observations to best explanations. B1 tests competitor inference (inferring a rival's cost structure from observed pricing). B2 tests strategy shift detection (identifying what changed from market outcome data). B3 tests hidden information extraction (inferring unobserved parameters from partial data).

Each question includes a closed-form or qualitatively definite expected answer, a grading rubric specifying key fields, and explicit scoring criteria. Questions are ordered by empirically observed difficulty (Section~\ref{sec:baselines}), spanning nearly the full 0--10 range.

\subsection{Intervention Design}

We test five conditions per question. The baseline presents the question with no scaffolding. The commitment condition asks the model to first state and commit to its analytical framework before solving, forcing explicit premise-locking. The contradiction detection condition prompts the model to identify and resolve internal contradictions in its reasoning chain. The principled separation condition requires solving in three explicit phases: state principles, generate predictions, then decide, imposing a declare-then-execute pipeline. The adversarial stress-test condition prompts the model to argue against its own conclusions and then defend or revise them.

\subsection{Prompt Framing, Repetition, and Evaluation}

Each question is presented in three framings: formal (mathematical notation and game-theoretic terminology), narrative (concrete scenario with natural language), and minimal (bare parameters and direct questions). Each combination of model, intervention, and framing is run 3 times at temperature~1.0, yielding $2 \times 5 \times 3 \times 3 \times 8 = 720$ responses. Temperature~1.0 was chosen to maximize response diversity across repetitions, providing a stronger test of intervention robustness than near-deterministic sampling. All responses are collected via the OpenAI API with a 32,768 completion token budget to avoid truncating reasoning model chain-of-thought.

Each response is scored by GPT-5.2 on two independent 0--10 scales: a conclusion score measuring factual accuracy of final outputs, and a reasoning score measuring derivation chain quality. We report the combined score as their arithmetic mean. For Component~B, the judge additionally evaluates declarative correctness (did the model identify the right principle?), procedural correctness (did it execute correctly?), and the gap between them.

To calibrate the automated judge, one author independently scored a stratified random sample of 180 responses (25\% of the 720 total), drawn proportionally across models, interventions, and questions. Of these 180, 176 received identical scores from the human and automated judge; the remaining 4 exhibited minor discrepancies of $\pm 1$ point on the 0--10 scale. This yields quadratic weighted Cohen's $\kappa = 0.97$, indicating near-perfect agreement. We report means with sample standard deviations; per-cell sample sizes are $n = 9$ (3 framings $\times$ 3 repetitions). Statistical tests use paired $t$-tests across the 8 per-question deltas (df $= 7$), supplemented by permutation tests and Wilcoxon signed-rank tests for robustness.

% ===================================================================
\section{Results}
\label{sec:results}
% ===================================================================

\subsection{Baseline Performance}
\label{sec:baselines}

Table~\ref{tab:baseline} reports baseline performance by question. Both models agree on a consistent difficulty hierarchy spanning nearly the full 0--10 scale.

\begin{table}
\centering
\small
\caption{Baseline combined scores by question (mean $\pm$ std, $n{=}9$).}
\label{tab:baseline}
\begin{tabular}{llccc}
\toprule
Question & Type & GPT-4.1-mini & GPT-5-mini & $\Delta$ \\
\midrule
A1: Param.\ Sensitivity & Deductive & $9.44 \pm 0.39$ & $9.61 \pm 0.22$ & $+0.17$ \\
A2: Location Change & Deductive & $5.89 \pm 0.49$ & $7.61 \pm 1.22$ & $+1.72$ \\
A3: Cost Structure ID & Deductive & $8.61 \pm 0.78$ & $4.22 \pm 1.09$ & $-4.39$ \\
A4: Asymmetric Costs & Deductive & $6.28 \pm 1.91$ & $7.33 \pm 1.54$ & $+1.05$ \\
A5: Entry Deterrence & Deductive & $2.53 \pm 0.43$ & $4.61 \pm 1.02$ & $+2.08$ \\
\midrule
B1: Competitor Inference & Abductive & $5.61 \pm 0.70$ & $5.94 \pm 0.17$ & $+0.33$ \\
B2: Strategy Shift & Abductive & $7.72 \pm 0.57$ & $7.83 \pm 0.75$ & $+0.11$ \\
B3: Hidden Information & Abductive & $3.94 \pm 0.73$ & $5.28 \pm 1.99$ & $+1.34$ \\
\midrule
Component A avg & & $6.55$ & $6.68$ & $+0.13$ \\
Component B avg & & $5.76$ & $6.35$ & $+0.59$ \\
Overall avg & & $6.25$ & $6.55$ & $+0.30$ \\
\bottomrule
\end{tabular}
\end{table}

GPT-5-mini outperforms GPT-4.1-mini on 6 of 8 questions, with the largest advantages on the hardest problems: A5 ($+2.08$), A2 ($+1.72$), and B3 ($+1.34$). The difficulty ordering is remarkably consistent across models: both rank A1 as easiest and A5 as hardest, with Spearman's $\rho = 0.62$ for the shared difficulty ranking. GPT-4.1-mini's sole notable advantage is A3, where it scores 8.61 versus 4.22---a 4.39-point reversal analyzed in Section~\ref{sec:a3}.

\subsection{The Crossover Interaction}

Table~\ref{tab:interventions} presents the central finding. Intervention effects differ not just in magnitude but in direction between models, producing a statistically significant crossover.

\begin{table}
\centering
\small
\caption{Mean intervention delta by model, averaged across 8 questions. Negative values indicate degradation relative to baseline. The final row reports paired $t$-test $p$-values for the crossover interaction.}
\label{tab:interventions}
\begin{tabular}{lcccc}
\toprule
Model & Commitment & Contradiction & Separation & Stress-test \\
\midrule
GPT-4.1-mini & $+0.21$ & $-0.27$ & $-0.40$ & $-0.57$ \\
GPT-5-mini   & $-0.63$ & $-0.76$ & $+0.31$ & $-1.47$ \\
\midrule
$\Delta$ (5m $-$ 4.1m) & $-0.84$ & $-0.49$ & $+0.71$ & $-0.90$ \\
$p$ (paired $t$) & $.040^{*}$ & $.088^{\dagger}$ & $.002^{**}$ & $.038^{*}$ \\
\bottomrule
\end{tabular}
\end{table}

Commitment scaffolding produces a crossover interaction ($t(7) = -2.52$, $p = 0.040$, $d = -0.89$): GPT-4.1-mini improves by $+0.21$ on average (positive in 6/8 questions), while GPT-5-mini degrades by $-0.63$ (negative in 7/8 questions, with losses reaching $-1.73$). The standard model lacks an internal commitment mechanism, so explicit premise-locking provides useful structure; the reasoning model already performs internal deliberation, and forcing external commitment creates interference---paralleling the verbal overshadowing effect \citep{schooler1990verbal, wilson1991thinking}.

Principled separation produces the reverse crossover ($t(7) = 4.90$, $p = 0.002$, $d = 1.73$): GPT-4.1-mini degrades by $-0.40$ (negative on all 8 questions), while GPT-5-mini improves by $+0.31$ (positive in 6/8 questions, with gains up to $+1.00$). The reasoning model benefits from phase-gating that structures its deliberation into a declare-then-execute pipeline; the standard model, lacking the reasoning capacity to exploit this structure, incurs overhead without benefit. The combined $2\times 2$ interaction is highly significant ($t(7) = 4.79$, $p = 0.002$, $d = 1.69$), with 7/8 questions showing the predicted crossover direction ($p_{\text{perm}} < 0.02$).

Adversarial stress-testing degrades both models, with $2.6\times$ greater damage to the reasoning model ($-1.47$ vs.\ $-0.57$; crossover $t(7) = -2.56$, $p = 0.038$, $d = -0.90$). GPT-5-mini's degradation is individually highly significant ($t(7) = -5.32$, $p = 0.001$, $d = -1.88$), while GPT-4.1-mini's is directional but not significant ($p = 0.20$).

\begin{figure}
  \centering
  \includegraphics[width=0.75\linewidth]{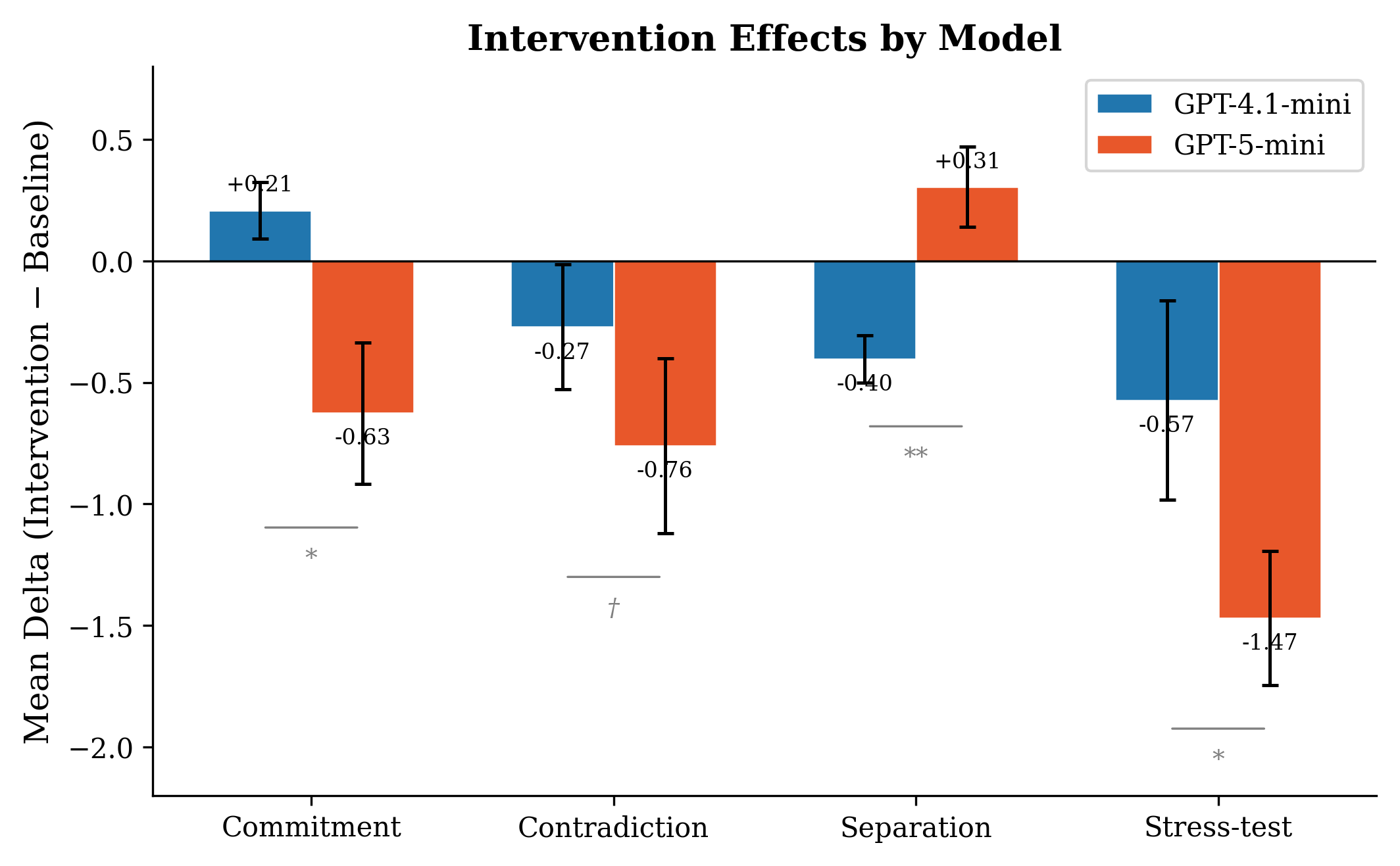}
  \caption{Crossover interaction between scaffolding type and model architecture. Commitment (leftmost pair) helps the standard model but hurts the reasoning model; separation (third pair) shows the reverse. Error bars show SEM across 8 per-question deltas. Brackets indicate paired interaction significance: $^{*}p < .05$; $^{**}p < .01$.}
  \label{fig:crossover}
\end{figure}

\subsection{Question-Level Analysis}
\label{sec:a3}

\begin{figure}
  \centering
  \includegraphics[width=0.75\linewidth]{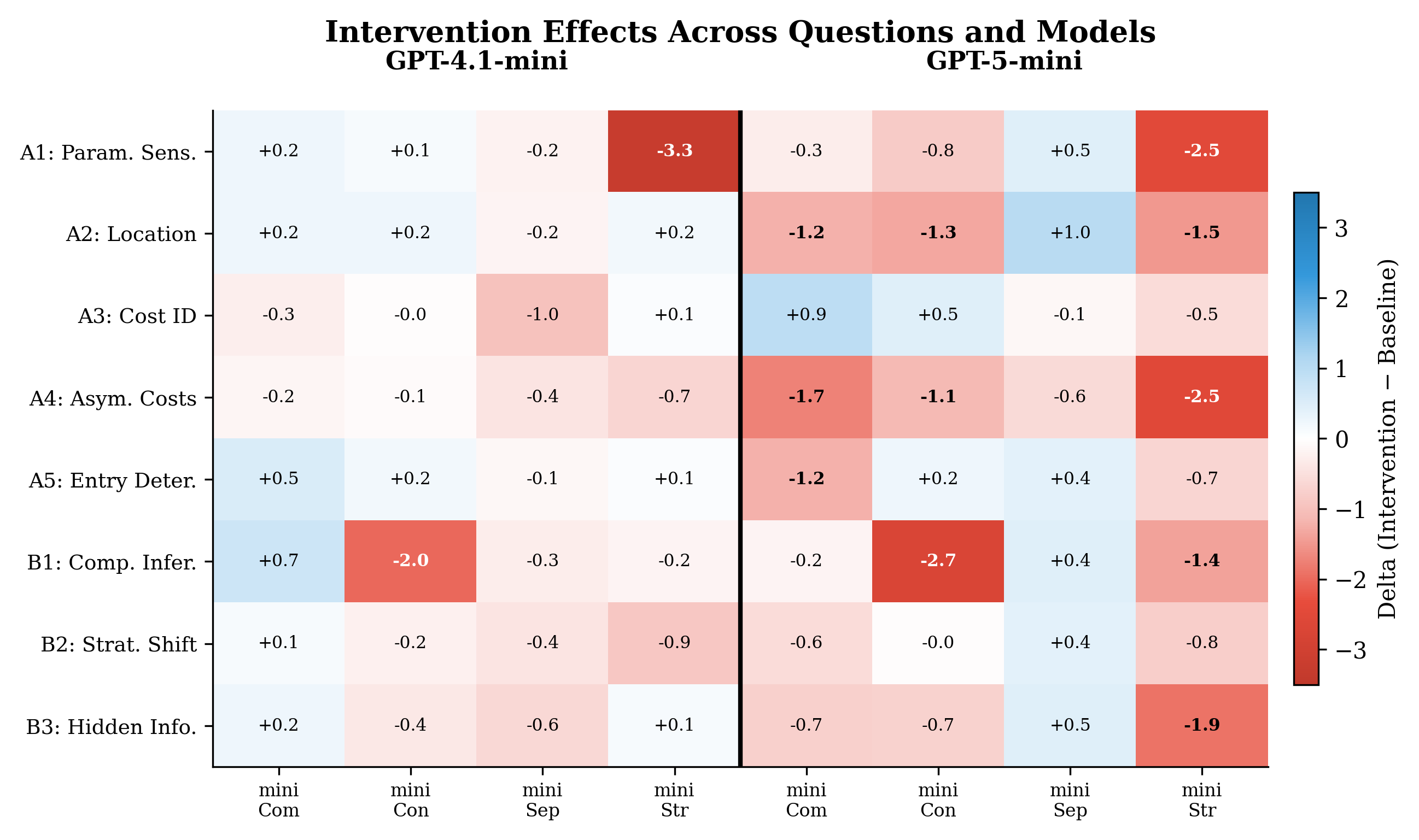}
  \caption{Intervention deltas across questions and models. Blue cells indicate improvement over baseline; red cells indicate degradation. The vertical black line separates the two models. Bold values exceed $\pm 1.0$.}
  \label{fig:heatmap}
\end{figure}

Table~\ref{tab:questions} reports the full question-level breakdown. Figure~\ref{fig:heatmap} visualizes the intervention deltas as a heatmap; the commitment columns show a clear blue-to-red gradient from GPT-4.1-mini to GPT-5-mini, with the pattern reversing for separation.

\begin{table}
\centering
\footnotesize
\setlength{\tabcolsep}{3.5pt}
\caption{Combined scores by question and intervention for both models.}
\label{tab:questions}
\begin{tabular}{ll|ccccc|ccccc}
\toprule
& & \multicolumn{5}{c|}{GPT-4.1-mini} & \multicolumn{5}{c}{GPT-5-mini} \\
Q & Task & Base & Com & Con & Sep & Str & Base & Com & Con & Sep & Str \\
\midrule
A1 & Param.  & 9.44 & 9.68 & 9.56 & 9.24 & 6.17 & 9.61 & 9.32 & 8.78 & 10.00 & 7.14 \\
A2 & Loc.    & 5.89 & 6.12 & 6.11 & 5.70 & 6.06 & 7.61 & 6.38 & 6.28 & 8.61 & 6.11 \\
A3 & Cost    & 8.61 & 8.34 & 8.56 & 7.65 & 8.67 & 4.22 & 5.16 & 4.67 & 4.11 & 3.67 \\
A4 & Asym.   & 6.28 & 6.12 & 6.22 & 5.85 & 5.60 & 7.33 & 5.60 & 6.22 & 6.75 & 4.81 \\
A5 & Entry   & 2.53 & 3.07 & 2.72 & 2.40 & 2.61 & 4.61 & 3.38 & 4.83 & 5.00 & 3.94 \\
\midrule
B1 & Infer.  & 5.61 & 6.34 & 3.61 & 5.32 & 5.44 & 5.94 & 5.77 & 3.22 & 6.38 & 4.56 \\
B2 & Shift   & 7.72 & 7.84 & 7.50 & 7.30 & 6.82 & 7.83 & 7.27 & 7.78 & 8.22 & 7.06 \\
B3 & Hidden  & 3.94 & 4.18 & 3.56 & 3.33 & 4.06 & 5.28 & 4.54 & 4.56 & 5.74 & 3.39 \\
\bottomrule
\end{tabular}
\end{table}

The heatmap confirms the crossover is broad and consistent: for commitment, GPT-4.1-mini shows positive or near-zero deltas on 6/8 questions while GPT-5-mini is negative on 7/8; for separation, the pattern reverses. This directional consistency across the full difficulty gradient strengthens the architectural interpretation.

The most significant individual result is A3, where GPT-4.1-mini scores 8.61 and GPT-5-mini scores 4.22---a 4.39-point reversal ($z = -2.31$). GPT-4.1-mini shows aligned conclusion and reasoning scores (8.78 and 8.44), while GPT-5-mini shows a 2.33-point inversion (reasoning 5.89 vs.\ conclusion 3.56): it performs the correct diagnostic but then classifies incorrectly, a textbook instance of computational split-brain syndrome \citep{zhang2025splitbrain}. A3 is the only question requiring empirical classification from data, and the reasoning model's extended chain-of-thought introduces branch points where the analysis can derail. Commitment improves GPT-5-mini's A3 score from 4.22 to 5.16---its largest positive commitment effect---consistent with early framework-locking preventing late-stage drift.

\subsection{Task-Level Heterogeneity}
\label{sec:heterogeneity}

The aggregate crossover interaction (Table~\ref{tab:interventions}) could in principle mask heterogeneous or reversed effects at the question level. We address this concern directly by examining per-question patterns.

\begin{figure}
  \centering
  \includegraphics[width=0.75\linewidth]{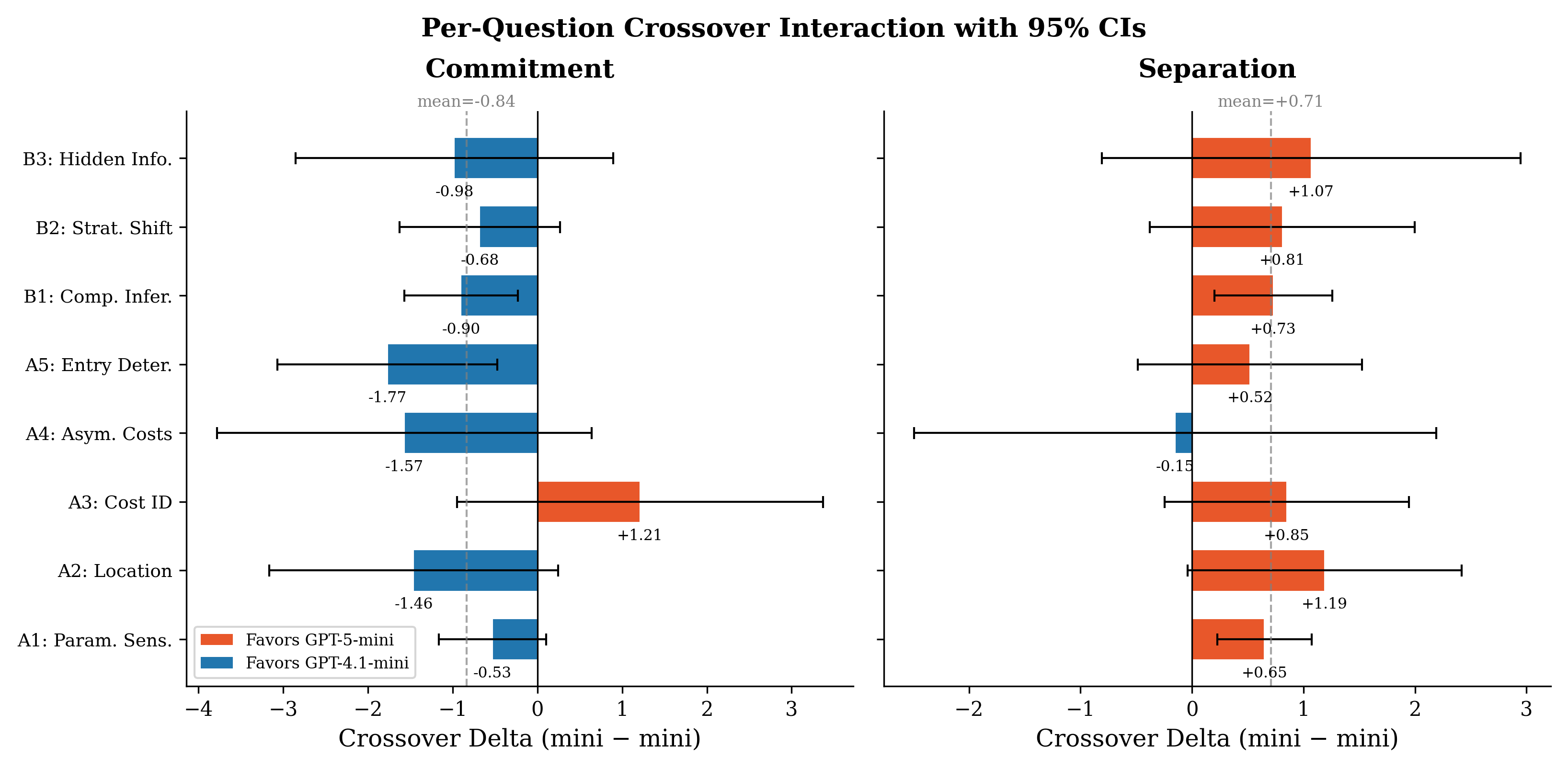}
  \caption{Per-question crossover deltas with 95\% confidence intervals for commitment (left) and separation (right). Orange bars favor GPT-5-mini; blue bars favor GPT-4.1-mini. The dashed line shows the mean crossover. Commitment shows 7/8 directional consistency with the A3 anomaly; separation shows 7/8 consistency with A4 as the sole exception.}
  \label{fig:forest}
\end{figure}

Figure~\ref{fig:forest} presents a forest plot of per-question crossover deltas with 95\% confidence intervals. For commitment, the crossover is negative in 7/8 questions, with A3 the sole exception. For separation, the crossover is positive in 7/8 questions. A4 (asymmetric cost analysis) is the only exception ($-0.15$): the sequential cost-comparison structure of this question---where pricing and location decisions are interdependent---resists clean phase decomposition, causing the three-phase pipeline to fragment rather than organize GPT-5-mini's reasoning.

The separation benefit shows a task-type gradient: abductive tasks (Component~B mean crossover: $+0.87$) benefit more than deductive tasks (Component~A mean crossover: $+0.61$). This is consistent with the separation protocol's three-phase pipeline (identify principles $\to$ generate predictions $\to$ decide), which maps directly onto the inference structure of abductive tasks where the model must first identify the relevant principle and then apply it to extract hidden information.

\begin{figure}
  \centering
  \includegraphics[width=0.75\linewidth]{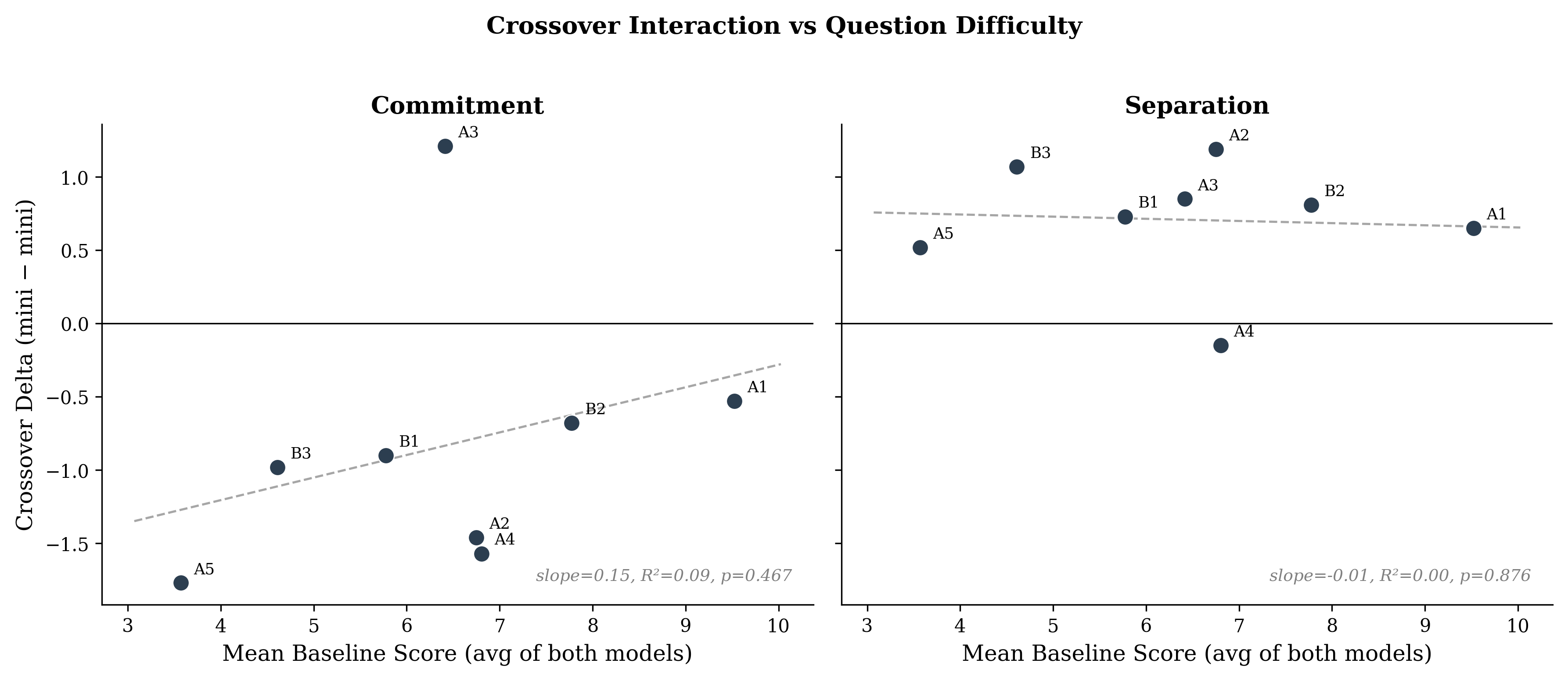}
  \caption{Crossover interaction magnitude vs.\ mean baseline difficulty. Neither intervention shows a significant difficulty gradient, indicating that the architecture$\times$scaffolding interaction operates uniformly across the difficulty spectrum.}
  \label{fig:difficulty}
\end{figure}

Figure~\ref{fig:difficulty} examines whether crossover magnitudes correlate with baseline difficulty. Neither commitment ($R^2 = 0.09$, $p = 0.47$) nor separation ($R^2 \approx 0$, $p = 0.93$) shows a significant difficulty gradient, confirming that the interaction effects are broadly distributed across the full difficulty spectrum rather than driven by easy or hard questions.

Excluding A3, the commitment crossover strengthens substantially ($t(6) = -6.29$, $p = 0.0008$) and the $2\times 2$ interaction reaches $p = 0.0001$. We retain A3 in all primary analyses to avoid post-hoc exclusion, but the sensitivity analysis confirms that the crossover findings are not dependent on this anomalous question.

\subsection{The Declarative--Procedural Gap}
\label{sec:dp}

\begin{figure}
  \centering
  \includegraphics[width=0.75\linewidth]{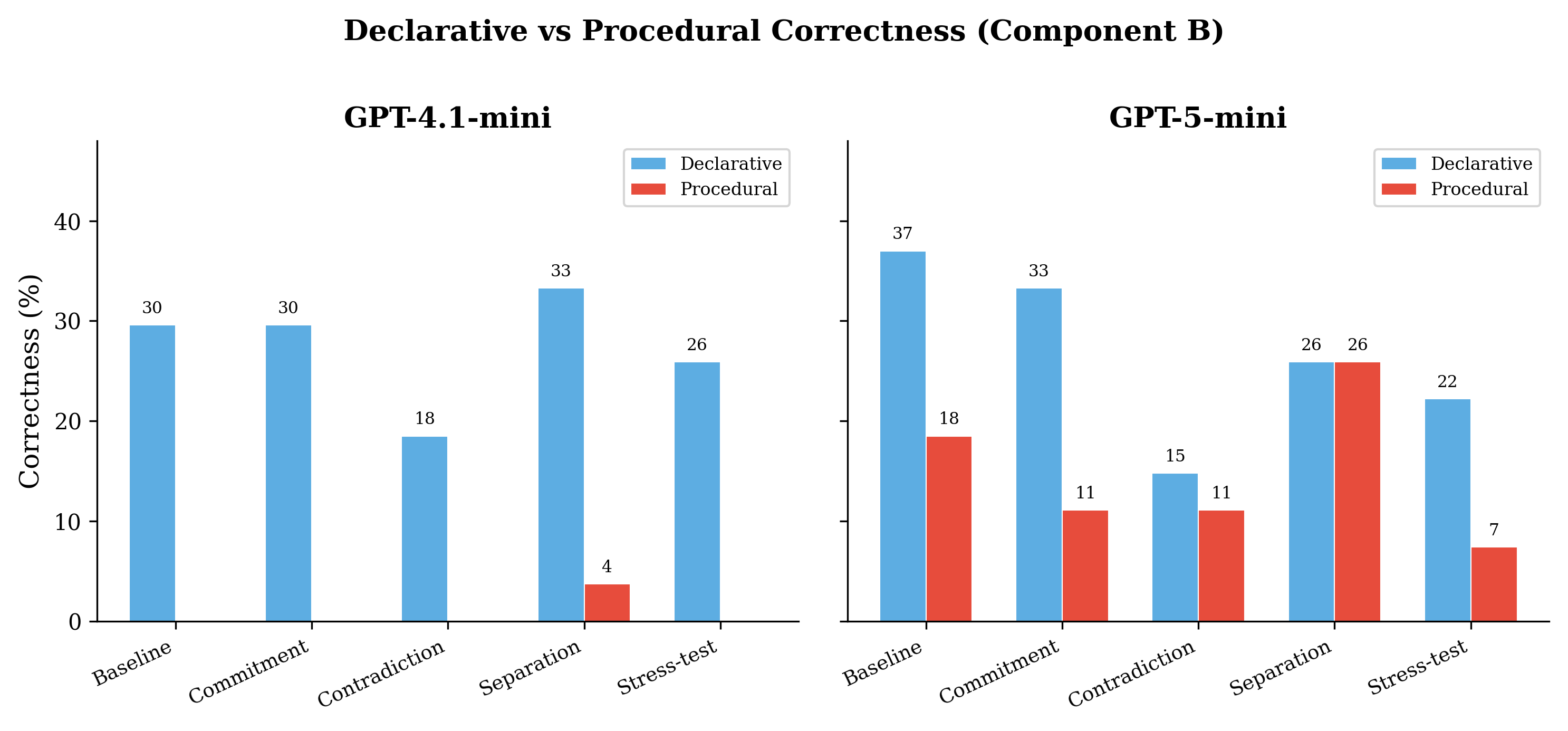}
  \caption{Declarative (blue) vs.\ procedural (red) correctness rates for Component~B. GPT-4.1-mini achieves near-zero procedural correctness regardless of intervention. GPT-5-mini under separation achieves convergence between declarative and procedural rates (both 25.9\%), fully closing the gap.}
  \label{fig:dp_gap}
\end{figure}

Table~\ref{tab:dp} reports the declarative--procedural gap for Component~B tasks, and Figure~\ref{fig:dp_gap} visualizes the pattern.

\begin{table}
\centering
\small
\caption{Declarative--procedural gap rates (\%) for Component~B across all conditions.}
\label{tab:dp}
\begin{tabular}{lccccc}
\toprule
& Baseline & Commit & Contra & Separate & Stress \\
\midrule
\multicolumn{6}{l}{\textit{GPT-4.1-mini}} \\
\quad Declarative & 29.6 & 29.6 & 18.5 & 33.3 & 25.9 \\
\quad Procedural & 0.0 & 0.0 & 0.0 & 3.7 & 0.0 \\
\quad Gap & 29.6 & 29.6 & 18.5 & 29.6 & 25.9 \\
\midrule
\multicolumn{6}{l}{\textit{GPT-5-mini}} \\
\quad Declarative & 37.0 & 33.3 & 14.8 & 25.9 & 22.2 \\
\quad Procedural & 18.5 & 11.1 & 11.1 & 25.9 & 7.4 \\
\quad Gap & 29.6 & 29.6 & 7.4 & 14.8 & 22.2 \\
\bottomrule
\end{tabular}
\end{table}

Across 135 Component~B judgments, GPT-4.1-mini translates declarative knowledge into correct action in exactly 1 instance (3.7\%, under separation only). The standard model describes correct strategies roughly 30\% of the time but implements them 0\% of the time, consistent with the Rylean distinction between knowing-that and knowing-how \citep{ryle1949concept}.

GPT-5-mini achieves meaningfully higher procedural rates, reaching 25.9\% under separation where declarative and procedural rates converge and the gap drops to zero. Under commitment, the procedural rate falls to 11.1\% while declarative holds at 33.3\%, widening the gap---further supporting the crossover finding.

Framing effects are small relative to intervention effects: the largest framing difference across all conditions is 0.64 points, compared to intervention deltas reaching $-1.73$ points, confirming that framing is secondary to the intervention$\times$architecture interaction.

% ===================================================================
\section{Discussion}
\label{sec:discussion}
% ===================================================================

\subsection{Architectural Interpretation}

The central finding is that scaffolding interventions interact with model architecture rather than operating uniformly. A standard model processes input in a single forward pass with no built-in deliberation; scaffolding supplies capabilities the model lacks, and commitment anchors an otherwise unconstrained generation process. A reasoning model already conducts internal chain-of-thought; duplicating this externally forces reconciliation between internal and imposed structure, degrading performance. When scaffolding instead complements existing capabilities---as separation does by organizing rather than duplicating deliberation---the model benefits. The design principle is: provide what the architecture lacks; do not duplicate what it already has.

That separation actively harms GPT-4.1-mini ($-0.40$, negative on all 8 questions) strengthens this interpretation: the model lacks the reasoning depth to exploit structured phase-gating, and the extra processing stage becomes pure overhead.

\subsection{The Stress-Test Paradox}

Adversarial self-examination deserves particular attention because it is the intervention most intuitively expected to improve robustness, yet it produces the worst outcomes. The effect is paradoxical: the largest damage occurs on the easiest problems, where models have already arrived at correct answers and are then induced to doubt them.

\begin{figure}
  \centering
  \includegraphics[width=0.75\linewidth]{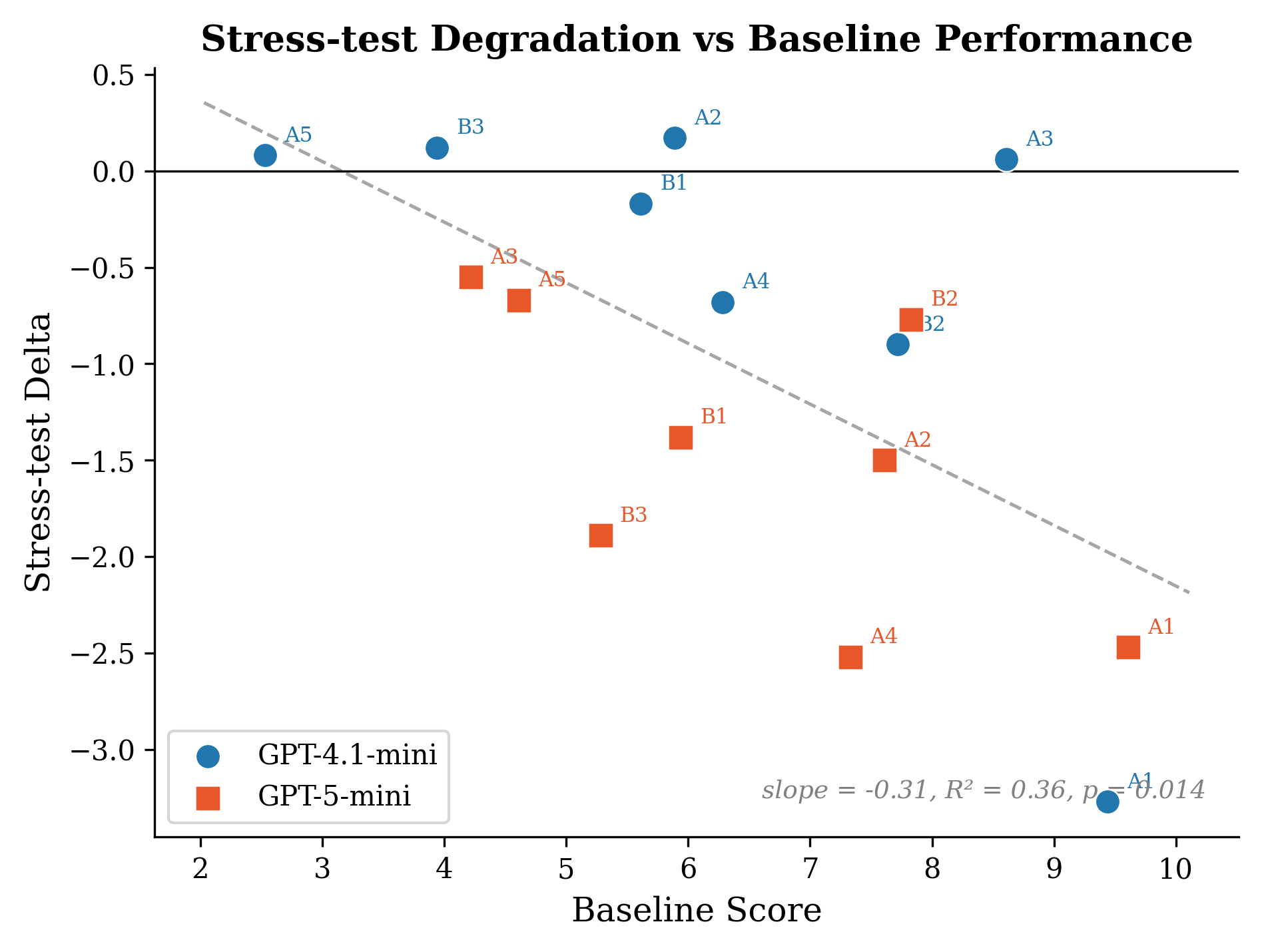}
  \caption{Stress-test degradation vs.\ baseline performance. Each point represents one model--question pair ($n = 16$). The negative slope indicates that well-understood problems suffer the greatest degradation under adversarial self-examination ($R^2 = 0.36$, $p = 0.014$).}
  \label{fig:stress_scatter}
\end{figure}

Baseline performance and stress-test degradation are negatively correlated (slope~$=-0.31$, $R^2=0.36$, $p=0.014$; Figure~\ref{fig:stress_scatter}): self-critique on correct answers provides surface area for doubt to propagate. The reasoning model's extended deliberation offers more such surface area, explaining the $2.6\times$ differential. This mirrors the "thinking too much" effect in human cognition, where articulating reasons for correct intuitive judgments degrades accuracy \citep{wilson1991thinking}.

\subsection{Diagnostic Value of the Framework}

The declarative--procedural gap provides the clearest window into architectural differences. Under separation, GPT-5-mini achieves perfect convergence between declarative and procedural rates at 25.9\%, completely closing the gap. Under commitment, the procedural rate drops to 11.1\% while declarative holds at 33.3\%, tripling the gap. For GPT-4.1-mini, no intervention produces procedural correctness above 3.7\%.

The gap appears to have fundamentally different origins in the two architectures. For the standard model, it reflects a capability limitation: the model cannot perform the multi-step inference required to translate principle to action. For the reasoning model, it reflects an organizational problem that scaffolding can resolve. This distinction has practical implications: for capability-limited models, scaffolding cannot substitute for reasoning capacity; for organizationally limited models, scaffolding choice matters as much as model choice.

More broadly, these results validate Hotelling as a diagnostic "microworld" \citep{papert1980mindstorms}: a structurally rich environment where intervention$\times$architecture interactions can be isolated from confounds.

\subsection{Limitations}

Our evaluation uses two proprietary models from a single provider (OpenAI), limiting both reproducibility and generalizability. While the GPT-4.1-mini and GPT-5-mini pair provides a clean architectural contrast---standard instruction-following versus reasoning-optimized---we cannot determine whether the observed interactions generalize to other provider families. The design principle itself (complement what the architecture lacks; do not duplicate what it already has) generates testable cross-provider hypotheses, and replication is a priority for future work. Per-cell sample sizes of $n = 9$ provide adequate power for the large effects we report ($d > 0.89$) but may miss smaller interactions. The LLM-as-judge methodology, despite $\kappa = 0.97$, may introduce response-style biases, though the 25\% human validation revealed no systematic bias. Our eight questions span a single economic domain; generalization to mechanism design, auction theory, and general equilibrium remains open.

% ===================================================================
\section{Conclusion}
\label{sec:conclusion}
% ===================================================================

We have demonstrated that reasoning interventions in LLMs do not have uniform effects, and that their impact depends critically and predictably on model architecture. Using Hotelling spatial competition as a diagnostic vehicle, we document a statistically significant crossover interaction ($p = 0.002$, $d = 1.69$) in which commitment scaffolding improves a standard model by $+0.21$ while degrading a reasoning model by $-0.63$, and principled separation improves the reasoning model by $+0.31$ while degrading the standard model by $-0.40$. Both individual crossovers are significant (commitment: $p = 0.040$; separation: $p = 0.002$), hold across 7/8 questions, and survive non-parametric robustness checks.

The declarative--procedural gap is fully closable by principled separation for the reasoning model but appears to be a fundamental capability limitation for the standard model. The adversarial stress-test paradox---where the easiest problems suffer the most---cautions against the intuition that self-criticism improves robustness. The largest per-question effects reach $-1.73$ for a mismatched intervention, exceeding the baseline model gap and motivating architecture-aware scaffolding.

% ===================================================================

\clearpage
\appendix
\section{Full Question Specifications}
\label{app:questions}

Each question is presented in three framings: formal, narrative, and minimal. All framings share the same JSON response schema. The narrative framing uses a "Shoreline Square" caf\'{e} scenario; the minimal framing strips economic context to pure mathematical parameters. Data tables referenced by \texttt{\{table\}} placeholders are shown inline.

\subsection*{Narrative Preamble (prepended to all narrative framings)}
\begin{small}
\begin{verbatim}
SETTING -- Shoreline Square
Shoreline Square is a 1 km * 1 km seaside resort district.  1,000 tourists
visit every day, spread evenly across the district.  Two rival cafes compete
for their business.  Each tourist buys from whichever cafe offers the lowest
*total cost* = menu price + walking hassle.  Walking hassle follows a
QUADRATIC rule: hassle = t * (straight-line distance in km)^2.  The hassle
parameter t captures how much tourists dislike walking (higher t = more
reluctance).  Unless stated otherwise, t = 1.0.

Cafe Alpha (yours) and Cafe Beta (the rival) each choose a location in the
square and a menu price.  All costs are per-unit; there are no fixed costs
unless stated otherwise.
\end{verbatim}
\end{small}

\subsection*{A1: Parameter Sensitivity (Component A)}
\paragraph{Formal Framing:}
\begin{small}
\begin{verbatim}
COUNTERFACTUAL REASONING: PARAMETER SENSITIVITY

BASELINE SCENARIO:
- Quadratic transport costs with parameter t = 1.0
- Two firms at opposite corners of the unit square: A at (0,0), B at (1,1)
- Equilibrium price = $25 each
- Market share = 50% each
- Profit per firm = $12,500 (market size = 1,000 consumers)

ASSUMPTIONS:
- Zero marginal cost for both firms
- Firms remain at opposite corners (maximal differentiation)
- Symmetric equilibrium: market shares stay 50% each
- Equilibrium price scales linearly with t: price = 25 * t

TASK: For each counterfactual value of t in {0.5, 2.0, 4.0}, derive the new
equilibrium price and profit per firm.
\end{verbatim}
\end{small}

\paragraph{Narrative Framing:}
\begin{small}
\begin{verbatim}
SCENARIO: WHAT IF TOURISTS WALK MORE (OR LESS) EASILY?

In Shoreline Square today (t = 1.0), both cafes sit at opposite corners of
the district: Cafe Alpha at the southwest corner (0, 0) and Cafe Beta at the
northeast corner (1, 1).  Each charges $25, splits the 1,000 daily tourists
50-50, and earns $12,500/day profit (zero ingredient cost).

The town council is debating three infrastructure proposals that would change
how much tourists mind walking:
  - Proposal 1 -- free shuttle bikes (t drops to 0.5): tourists barely
    mind distance
  - Proposal 2 -- remove benches, add hills (t rises to 2.0): walking is
    harder
  - Proposal 3 -- extreme heat wave (t rises to 4.0): tourists hate walking

Assume both cafes stay at their current corners and the market stays
symmetric (50-50 split).  Equilibrium price scales linearly: price = 25 * t.

TASK: For each proposal, derive the new menu price and daily profit per cafe.
\end{verbatim}
\end{small}

\paragraph{Minimal Framing:}
\begin{small}
\begin{verbatim}
OPTIMIZATION PROBLEM: PARAMETER SENSITIVITY

Two agents are fixed at positions (0,0) and (1,1) in [0,1]^2.  A population
of 1,000 points is uniformly distributed on the square.  Each point is
allocated to the agent offering the lowest total cost:
    C_i(x,y) = v_i + t * [(x - a_i)^2 + (y - b_i)^2]
where v_i is the value set by agent i at position (a_i, b_i).

At t = 1.0 the symmetric Nash equilibrium has v* = 25, each agent captures
500 points, and each agent's payoff = 12,500.

GIVEN: Equilibrium value scales linearly with t: v* = 25 * t.

TASK: Compute v* and payoff for t in {0.5, 2.0, 4.0}.
\end{verbatim}
\end{small}

\paragraph{JSON response schema.}
\begin{small}
\begin{verbatim}
Respond in JSON format:
{
    "baseline": {
        "t": 1.0, "price": 25, "profit_per_firm": 12500,
        "explanation": "baseline derivation"
    },
    "counterfactual_t_0.5": {
        "price": "<derived value>",
        "profit_per_firm": "<derived value>",
        "derivation": "step-by-step explanation"
    },
    "counterfactual_t_2.0": {
        "price": "<derived value>",
        "profit_per_firm": "<derived value>",
        "derivation": "step-by-step explanation"
    },
    "counterfactual_t_4.0": {
        "price": "<derived value>",
        "profit_per_firm": "<derived value>",
        "derivation": "step-by-step explanation"
    },
    "general_relationship": {
        "price_formula": "price as a function of t",
        "profit_formula": "profit as a function of t",
        "economic_intuition": "why this relationship holds"
    }
}
\end{verbatim}
\end{small}

\subsection*{A2: Location Change (Component A)}
\paragraph{Formal Framing:}
\begin{small}
\begin{verbatim}
COUNTERFACTUAL REASONING: LOCATION CHANGE

BASELINE:
- Quadratic transport costs, t = 1.0
- Firm A at (0,0), Firm B at (1,1); distance = sqrt2 ~= 1.414
- Equilibrium price = $25 each, market share = 50% each

COUNTERFACTUAL: Firm B relocates from (1,1) to (0.5, 0.5).  Firm A stays
at (0,0).  Both firms then re-optimize prices.

TASK: Derive the consequences of this relocation.
\end{verbatim}
\end{small}

\paragraph{Narrative Framing:}
\begin{small}
\begin{verbatim}
SCENARIO: CAFE BETA MOVES TO THE TOWN CENTER

Currently, Cafe Alpha sits at the southwest corner (0, 0) and Cafe Beta at
the northeast corner (1, 1) of Shoreline Square.  They are sqrt2 ~= 1.414 km
apart, each charges $25, splits tourists 50-50.

Cafe Beta just announced it will relocate to the exact center of the district
(0.5, 0.5).  Cafe Alpha is staying put.  Both will reset their menu prices
once the move is complete.

TASK: Analyze what happens to prices, market shares, and profits for both
cafes after Beta's move.
\end{verbatim}
\end{small}

\paragraph{Minimal Framing:}
\begin{small}
\begin{verbatim}
POSITION CHANGE ANALYSIS

Two agents in [0,1]^2.  Agent A at (0,0), Agent B at (1,1).
Cost: C_i(x,y) = v_i + t*[(x-a_i)^2+(y-b_i)^2], t = 1.0.
Baseline: v* = 25 each, shares 0.5 each, payoff 12,500 each.

Agent B moves to (0.5, 0.5).  Agent A stays at (0,0).  Both re-optimize v_i.

TASK: Derive the new inter-agent distance, direction of value/share/payoff
changes, and who benefits.
\end{verbatim}
\end{small}

\paragraph{JSON response schema (truncated).}
\begin{small}
\begin{verbatim}
Respond in JSON format:
{
    "baseline": {
        "distance": 1.414, "prices": [25, 25],
        "shares": [0.5, 0.5], "profits": [12500, 12500]
    },
    "new_configuration": {
        "firm_a_location": [0, 0], "firm_b_location": [0.5, 0.5],
        "new_distance": "<calculated>",
        "distance_change_percent": "<percentage change>"
    },
    "predicted_effects": {
        "competition_intensity": "increase or decrease",
        "equilibrium_prices": "increase or decrease",
        "price_estimate": "<approx new equilibrium price>",
        "firm_a_market_share": "increase or decrease or unchanged",
        "firm_b_market_share": "increase or decrease or unchanged"
    },
    "derivation": [
        {"step": 1, "claim": "...", "reasoning": "..."},
        {"step": 2, "claim": "...", "reasoning": "..."},
        {"step": 3, "claim": "...", "reasoning": "..."}
    ],
    "who_benefits": "Firm A or Firm B or neither or both harmed",
    "profit_analysis": {
    ... [remaining fields omitted for space]
\end{verbatim}
\end{small}

\subsection*{A3: Cost Structure Identification (Component A)}
Data table (embedded in all framings):
\begin{small}
\begin{verbatim}
| Config | Distance | Price_A | Price_B | Share_A | Share_B |
|--------|----------|---------|---------|---------|---------|
| 1      | 0.50     | 30      | 30      | 0.500   | 0.500   |
| 2      | 0.50     | 29      | 30      | 0.731   | 0.269   |
| 3      | 1.00     | 30      | 30      | 0.500   | 0.500   |
| 4      | 1.00     | 29      | 30      | 0.608   | 0.392   |
\end{verbatim}
\end{small}

\paragraph{Formal Framing:}
\begin{small}
\begin{verbatim}
BACKWARD DEDUCTION: COST STRUCTURE IDENTIFICATION

Two firms compete across four experimental configurations.
The transport cost structure (LINEAR vs QUADRATIC) is unknown.

OBSERVED DATA:
[DATA TABLE ABOVE]
[HINT TEXT]
\end{verbatim}
\end{small}

\paragraph{Narrative Framing:}
\begin{small}
\begin{verbatim}
SCENARIO: WHAT KIND OF WALKING HASSLE?

Shoreline Square's tourism board ran an experiment with two pop-up cafes at
varying separations.  In some trials one cafe offered a $1 discount.  You
have the results but you do NOT know whether walking hassle is proportional
to distance (linear) or to distance-squared (quadratic).

EXPERIMENTAL RESULTS:
[DATA TABLE ABOVE]
[HINT TEXT]
\end{verbatim}
\end{small}

\paragraph{Minimal Framing:}
\begin{small}
\begin{verbatim}
MODEL SELECTION FROM ALLOCATION DATA

Two agents compete; the cost function is EITHER  C = v + t*d  (linear) OR
C = v + t*d^2  (quadratic), where d is Euclidean distance.  You must infer
which model generated the following data:

[DATA TABLE ABOVE]
[HINT TEXT]
\end{verbatim}
\end{small}

\paragraph{JSON response schema.}
\begin{small}
\begin{verbatim}
Respond in JSON format:
{
    "analysis": {
        "config_2_share_gain": "<Share_A increase from equal prices>",
        "config_4_share_gain": "<Share_A increase from equal prices>",
        "pattern_observed": "description of key pattern"
    },
    "cost_structure_determination": {
        "conclusion": "LINEAR or QUADRATIC",
        "diagnostic_evidence": "which comparison reveals the structure",
        "reasoning": "step-by-step explanation"
    },
    "parameter_estimation": {
        "estimated_t": "<number>",
        "estimation_method": "how you estimated t",
        "confidence": "high or medium or low"
    },
    "verification": {
        "does_data_fit": "true or false",
        "consistency_check": "verify against another data point"
    }
}
\end{verbatim}
\end{small}

\subsection*{A4: Asymmetric Cost Analysis (Component A)}
\paragraph{Formal Framing:}
\begin{small}
\begin{verbatim}
COUNTERFACTUAL REASONING: ASYMMETRIC COSTS

BASELINE: Quadratic transport costs, t = 1.0.  Both firms at opposite
corners, marginal cost = $0, price = $25, share = 50%, profit = $12,500.

COUNTERFACTUAL: Firm A's marginal cost rises to $10/unit.  Firm B stays
at $0.  Firms may adjust both price and location.

TASK -- answer in order:
1. If A keeps price = $25, what is its margin per sale?
2. If A raises price to $35 (restoring $25 margin) while B stays at $25,
   what happens?
3. Should A change its LOCATION in response to the cost disadvantage?
4. What is the qualitative new equilibrium?
\end{verbatim}
\end{small}

\paragraph{Narrative Framing:}
\begin{small}
\begin{verbatim}
SCENARIO: CAFE ALPHA'S COSTS SPIKE

Cafe Alpha just signed a contract with a high-end organic supplier: ingredient
costs jump to $10 per drink.  Cafe Beta still sources cheaply at $0.

Previously both cafes charged $25 at opposite corners, splitting tourists
50-50 for $12,500/day each.

TASK -- answer in order:
1. If Alpha keeps the $25 menu price, what is its profit margin per drink?
2. If Alpha raises to $35 (restoring $25 margin) while Beta stays at $25,
   what happens to market shares?
3. Should Alpha MOVE to a different location because of its cost disadvantage?
4. Describe the new equilibrium qualitatively.
\end{verbatim}
\end{small}

\paragraph{Minimal Framing:}
\begin{small}
\begin{verbatim}
ASYMMETRIC COST ANALYSIS

Baseline: agents at (0,0) and (1,1), v* = 25, payoff 12,500 each, marginal
cost = 0 for both.  Allocation: C_i = v_i + t*d^2, t = 1.0.

Change: Agent A's marginal cost rises to 10.  Agent B stays at 0.
Agents may adjust both v_i and position.

TASK:
1. A's per-unit margin if v_A stays 25?
2. Effect of A raising v_A to 35 while B holds v_B = 25?
3. Should A change its position? Which direction?
4. Qualitative equilibrium characterization.
\end{verbatim}
\end{small}

\paragraph{JSON response schema (truncated).}
\begin{small}
\begin{verbatim}
Respond in JSON format:
{
    "immediate_impact": {
        "firm_a_margin_at_old_price": "<profit per unit if price stays $25>",
        "firm_a_viability": "viable or squeezed or negative margin",
        "explanation": "why this margin level is problematic or acceptable"
    },
    "price_adjustment_scenario": {
        "if_a_raises_to_35": {
            "price_gap": "<B's price advantage>",
            "share_shift_direction": "toward A or toward B",
            "firm_a_tradeoff": "description of margin vs volume tradeoff"
        }
    },
    "location_response": {
        "should_a_relocate": "true or false",
        "optimal_direction": "toward center or stay at corner or toward B",
        "reasoning": "why location change does or doesn't help"
    },
    "equilibrium_analysis": {
        "firm_a_equilibrium_price": "higher than B or equal to B or
                                     lower than B",
        "firm_a_equilibrium_share": "above 50% or equal 50% or below 50%",
        "firm_a_equilibrium_profit": "above B or equal to B or below B",
        "qualitative_outcome": "description of the new equilibrium"
    },
    ... [remaining fields omitted for space]
\end{verbatim}
\end{small}

\subsection*{A5: Entry Deterrence (Component A)}
\paragraph{Formal Framing:}
\begin{small}
\begin{verbatim}
SEQUENTIAL STRATEGIC REASONING: ENTRY DETERRENCE

SCENARIO:
- Quadratic transport costs, t = 1.0
- Market: [0,1]^2 with 1,000 uniformly distributed consumers
- Firm A (incumbent) at center (0.5, 0.5), charging $40, capturing all
  consumers
- Firm B (potential entrant) deciding whether and where to enter
- Both firms: marginal cost = $0
- Entry fixed cost for B = $5,000
- Current incumbent profit = $40,000

PART 1: If B enters at (0, 0), what is the post-entry equilibrium?
PART 2: Compare B entering at (0, 0) vs co-locating at (0.5, 0.5).
PART 3: What is B's optimal entry location given A is at center?
PART 4: Could A have chosen a different initial location to deter entry?
\end{verbatim}
\end{small}

\paragraph{Narrative Framing:}
\begin{small}
\begin{verbatim}
SCENARIO: A NEW CAFE MIGHT OPEN

Cafe Alpha is the only cafe in Shoreline Square, sitting right in the center
(0.5, 0.5) and charging $40 per drink.  It serves all 1,000 daily tourists
for $40,000/day profit (zero ingredient cost).

A rival entrepreneur is considering opening Cafe Beta.  Opening requires a
$5,000/day fixed cost (rent, equipment, staff).  Both cafes would have $0
ingredient costs.  The entrepreneur must decide: enter or not, and if so,
where?

PART 1: If Cafe Beta opens at the southwest corner (0, 0), what happens to
prices, shares, and profits?
PART 2: Compare opening at the corner vs opening right next to Alpha at
(0.5, 0.5).
PART 3: Where should Cafe Beta open to maximize profit?
PART 4: Could Cafe Alpha have chosen a different location initially to
discourage entry?
\end{verbatim}
\end{small}

\paragraph{Minimal Framing:}
\begin{small}
\begin{verbatim}
SEQUENTIAL ENTRY GAME

Agent A (incumbent) at (0.5, 0.5) in [0,1]^2, v_A = 40, capturing all 1,000
points.  Payoff = 40,000.  Agent B may enter at any position; entry costs
F = 5,000.  Both agents: marginal cost 0.  Post-entry, both set values v_i
simultaneously.  C_i(x,y) = v_i + 1.0*[(x-a_i)^2+(y-b_i)^2].

PART 1: Post-entry equilibrium if B enters at (0, 0)?
PART 2: Compare B at (0, 0) vs B at (0.5, 0.5).
PART 3: B's optimal entry position?
PART 4: Can A choose an initial position to deter entry?
\end{verbatim}
\end{small}

\paragraph{JSON response schema (truncated).}
\begin{small}
\begin{verbatim}
Respond in JSON format:
{
    "part_1_corner_entry": {
        "post_entry_equilibrium": {
            "firm_a_price": "<estimated>", "firm_b_price": "<estimated>",
            "firm_a_share": "<estimated>", "firm_b_share": "<estimated>"
        },
        "firm_b_gross_profit": "<before entry cost>",
        "firm_b_net_profit": "<after $5000 entry cost>",
        "entry_profitable": "true or false",
        "reasoning": "explanation"
    },
    "part_2_location_comparison": {
        "corner_entry": {
            "competition_intensity": "high or medium or low",
            "expected_prices": "high or medium or low",
            "b_profit_potential": "high or medium or low"
        },
        "colocation_entry": {
            "competition_intensity": "high or medium or low",
            "expected_prices": "high or medium or low",
            "b_profit_potential": "high or medium or low"
        },
        "better_entry_location": "corner or colocation",
        "reasoning": "why one dominates"
    ... [remaining fields omitted for space]
\end{verbatim}
\end{small}

\subsection*{B1: Competitor Inference (Component B)}
Observed data (embedded in all framings):
\begin{small}
\begin{verbatim}
Round  | Location      | Price
-------|---------------|-------
1      | (0.50, 0.50)  | $30
2      | (0.55, 0.55)  | $31
3      | (0.60, 0.60)  | $32
4      | (0.65, 0.65)  | $34
5      | (0.72, 0.72)  | $36
6      | (0.78, 0.78)  | $38
7      | (0.84, 0.84)  | $40
8      | (0.90, 0.90)  | $42
9      | (0.95, 0.95)  | $44
10     | (0.98, 0.98)  | $45
\end{verbatim}
\end{small}

\paragraph{Formal Framing:}
\begin{small}
\begin{verbatim}
BEHAVIORAL INFERENCE: COMPETITOR STRATEGY

You observe a competitor over 10 rounds in a market with quadratic transport
costs (t = 1.0) on [0,1]^2.  You are located at (0.1, 0.1) throughout.

OBSERVED COMPETITOR BEHAVIOR:
[DATA TABLE ABOVE]

TASK: Infer the competitor's strategy and predict future behavior.
\end{verbatim}
\end{small}

\paragraph{Narrative Framing:}
\begin{small}
\begin{verbatim}
SCENARIO: WATCHING CAFE BETA'S MOVES

You run Cafe Alpha at position (0.1, 0.1) in Shoreline Square--near the
southwest corner.  Over the past 10 days you've tracked Cafe Beta's location
and menu price.  (Quadratic walking hassle, t = 1.0.)

CAFE BETA'S DAILY LOG:
[DATA TABLE ABOVE]

TASK: Figure out what Cafe Beta is doing, predict where it's headed, and
decide your best response.
\end{verbatim}
\end{small}

\paragraph{Minimal Framing:}
\begin{small}
\begin{verbatim}
STRATEGY INFERENCE FROM SEQUENTIAL OBSERVATIONS

Agent B's position and value over 10 periods in [0,1]^2, t = 1.0.
Your position: fixed at (0.1, 0.1).

[DATA TABLE ABOVE]

TASK: Infer Agent B's objective, predict future (position, value), and
determine your optimal response.
\end{verbatim}
\end{small}

\paragraph{JSON response schema (truncated).}
\begin{small}
\begin{verbatim}
Respond in JSON format:
{
    "pattern_analysis": {
        "location_pattern": "description of how location changes over rounds",
        "location_velocity": "approximate movement per round",
        "price_pattern": "description of how price changes over rounds",
        "price_velocity": "approximate price change per round",
        "correlation": "relationship between location and price changes"
    },
    "candidate_strategies": [
        {
            "id": "S1", "name": "strategy name",
            "description": "what the competitor is doing",
            "objective": "what they are trying to achieve",
            "fit_to_data": "how well this explains the observations"
        },
        {
            "id": "S2", "name": "strategy name",
            "description": "what the competitor is doing",
            "objective": "what they are trying to achieve",
            "fit_to_data": "how well this explains the observations"
        }
    ],
    "best_explanation": "S1 or S2",
    "inferred_goal": "the competitor's ultimate objective in one sentence",
    ... [remaining fields omitted for space]
\end{verbatim}
\end{small}

\subsection*{B2: Strategy Shift Detection (Component B)}
Observed data (embedded in all framings):
\begin{small}
\begin{verbatim}
Round | Location      | Price  | Your Share
------|---------------|--------|----------
1     | (0.50, 0.50)  | $35    | 0.38
2     | (0.55, 0.55)  | $36    | 0.39
3     | (0.60, 0.60)  | $37    | 0.40
4     | (0.65, 0.65)  | $38    | 0.42
5     | (0.70, 0.70)  | $39    | 0.43
6     | (0.68, 0.68)  | $34    | 0.35
7     | (0.60, 0.60)  | $30    | 0.32
8     | (0.52, 0.52)  | $26    | 0.30
9     | (0.48, 0.48)  | $24    | 0.28
10    | (0.45, 0.45)  | $23    | 0.27
11    | (0.42, 0.42)  | $22    | 0.26
12    | (0.40, 0.40)  | $21    | 0.25
\end{verbatim}
\end{small}

\paragraph{Formal Framing:}
\begin{small}
\begin{verbatim}
BEHAVIORAL INFERENCE: STRATEGY SHIFT DETECTION

You observe your competitor over 12 rounds.  Quadratic transport costs,
t = 1.0.  Your location throughout: (0.1, 0.1).  Your price throughout: $35.

OBSERVED COMPETITOR BEHAVIOR:
[DATA TABLE ABOVE]

TASK: Detect and analyze the strategy shift.
\end{verbatim}
\end{small}

\paragraph{Narrative Framing:}
\begin{small}
\begin{verbatim}
SCENARIO: CAFE BETA SUDDENLY CHANGES TACTICS

You run Cafe Alpha at (0.1, 0.1), charging a steady $35/drink.  For 12 days
you've tracked Cafe Beta.  Something changed mid-period.

CAFE BETA'S DAILY LOG:
[DATA TABLE ABOVE]

TASK: Identify when and why Cafe Beta changed strategy, predict what comes
next, and decide how you should respond.
\end{verbatim}
\end{small}

\paragraph{Minimal Framing:}
\begin{small}
\begin{verbatim}
CHANGE-POINT DETECTION IN SEQUENTIAL STRATEGY DATA

Your position: (0.1, 0.1), your value: 35 (fixed).  Agent B's behavior:

[DATA TABLE ABOVE]

TASK: Detect the strategy change point, characterize both phases, predict
future behavior, and determine your optimal response.
\end{verbatim}
\end{small}

\paragraph{JSON response schema (truncated).}
\begin{small}
\begin{verbatim}
Respond in JSON format:
{
    "shift_detection": {
        "shift_occurred": "true or false",
        "shift_round": "<round number>",
        "confidence": "high or medium or low",
        "detection_method": "how you identified the shift"
    },
    "phase_1_analysis": {
        "rounds": "1 to X",
        "location_trend": "description",
        "price_trend": "description",
        "strategy_name": "name for this strategy",
        "strategy_description": "what competitor was doing",
        "apparent_objective": "what they were trying to achieve"
    },
    "phase_2_analysis": {
        "rounds": "X to 12",
        "location_trend": "description",
        "price_trend": "description",
        "strategy_name": "name for this strategy",
        "strategy_description": "what competitor is now doing",
        "apparent_objective": "what they are now trying to achieve"
    },
    "shift_hypotheses": [
    ... [remaining fields omitted for space]
\end{verbatim}
\end{small}

\subsection*{B3: Hidden Information Extraction (Component B)}
Observed data (embedded in all framings):
\begin{small}
\begin{verbatim}
Round | Your Location | Your Price | Your Market Share
------|---------------|------------|------------------
1     | (0.20, 0.20)  | $30        | 0.42
2     | (0.20, 0.20)  | $28        | 0.48
3     | (0.20, 0.20)  | $25        | 0.55
4     | (0.30, 0.30)  | $25        | 0.48
5     | (0.40, 0.40)  | $25        | 0.41
6     | (0.40, 0.40)  | $22        | 0.49
\end{verbatim}
\end{small}

\paragraph{Formal Framing:}
\begin{small}
\begin{verbatim}
BEHAVIORAL INFERENCE: HIDDEN INFORMATION

You can observe ONLY your own actions and resulting market share.  You CANNOT
directly observe the competitor's location or price.

Market: [0,1]^2, quadratic transport costs, t = 1.0, 1,000 consumers.

YOUR OBSERVATIONS:
[DATA TABLE ABOVE]

TASK: Infer the competitor's hidden state and determine your optimal action.
\end{verbatim}
\end{small}

\paragraph{Narrative Framing:}
\begin{small}
\begin{verbatim}
SCENARIO: THE INVISIBLE RIVAL

You run Cafe Alpha in Shoreline Square.  A rival cafe exists somewhere in
the district, but you don't know its location or menu price -- you only see
how many tourists come to YOUR cafe each day.

Over 6 days you've experimented with your own location and pricing:

YOUR DAILY LOG:
[DATA TABLE ABOVE]

TASK: From your own sales data, figure out where the rival is, what they're
charging, and decide your next move.
\end{verbatim}
\end{small}

\paragraph{Minimal Framing:}
\begin{small}
\begin{verbatim}
INVERSE INFERENCE FROM OWN-OUTCOME DATA

You observe only your own position, value, and resulting allocation share.
The competing agent's position and value are hidden.

Domain: [0,1]^2, C_i = v_i + 1.0*d^2, population = 1,000.

YOUR DATA:
[DATA TABLE ABOVE]

TASK: Infer the hidden agent's position and value.  Determine your optimal
action for period 7.
\end{verbatim}
\end{small}

\paragraph{JSON response schema (truncated).}
\begin{small}
\begin{verbatim}
Respond in JSON format:
{
    "inference_from_price_variation": {
        "rounds_compared": "1 vs 2 vs 3",
        "observation": "how share changed when you changed price",
        "inference": "what this reveals about competitor's price",
        "estimated_competitor_price": "<number or range>",
        "confidence": "high or medium or low",
        "reasoning": "step-by-step logic"
    },
    "inference_from_location_variation": {
        "rounds_compared": "3 vs 4 vs 5",
        "observation": "how share changed when you moved",
        "inference": "what this reveals about competitor's location",
        "estimated_competitor_location": "[x, y] or range",
        "confidence": "high or medium or low",
        "reasoning": "step-by-step logic"
    },
    "combined_inference": {
        "competitor_location_estimate": "[x, y]",
        "competitor_price_estimate": "<number>",
        "consistency_check": "do estimates explain all observations?",
        "remaining_uncertainty": "what you still don't know"
    },
    "round_6_analysis": {
    ... [remaining fields omitted for space]
\end{verbatim}
\end{small}

\section{Intervention Prompt Templates}
\label{app:interventions}

Each intervention modifies the base question prompt. We show the complete prompt template for each intervention type.

\subsection*{Commitment Protocol}
The following preamble is prepended to the base question:

\begin{small}
\begin{verbatim}
=== COMMITMENT PROTOCOL ===

Before answering this question you MUST:
1. State 1-3 PRINCIPLES that govern your reasoning.
2. These principles become BINDING COMMITMENTS -- your answer must follow
   from them.
3. All future answers in this session must remain consistent with every
   prior commitment.

After stating your principles, answer the question. If your natural answer
would violate a prior commitment, you must EITHER revise the answer OR
explicitly argue why the commitment should be updated.

Include your commitments in your JSON response under a top-level
"commitments" key:
  "commitments": [
    {"id": 1, "principle": "..."},
    {"id": 2, "principle": "..."}
  ]

Then provide the rest of your answer in the format requested by the question.
\end{verbatim}
\end{small}

For subsequent questions, inherited commitments are injected:

\begin{small}
\begin{verbatim}
=== INHERITED COMMITMENTS ===
You previously established these binding commitments:
{commitments}

Your answer to this question MUST be consistent with ALL inherited
commitments.  You may NOT silently contradict an inherited commitment.
\end{verbatim}
\end{small}

\subsection*{Principled Separation Protocol}
The base question is wrapped inside the following template (\texttt{{question}} is replaced with the original prompt):

\begin{small}
\begin{verbatim}
=== PRINCIPLE-PREDICTION-DECISION SEPARATION PROTOCOL ===

You MUST answer in three strictly separated phases. Complete each phase
fully BEFORE moving to the next. Do NOT look ahead to the question's
specifics while writing Phase 1.

PHASE 1 -- PRINCIPLES (general theory, no numbers from this problem)

State 2-3 ABSTRACT, FALSIFIABLE, CAUSAL principles that govern this
class of problem. Each principle must:
  1. Be stated in general terms (no specific values from the question).
  2. Identify a causal mechanism ("X causes Y because Z").
  3. Be classifiable as one of: equilibrium, comparative-static,
     information-theoretic, or strategic-interaction.

PHASE 2 -- PREDICTIONS (apply principles to THIS specific problem)

For each prediction you make:
  1. CITE which principle(s) it derives from by ID (e.g., P1, P2).
  2. Show the LOGICAL DERIVATION -- the chain of reasoning from principle
     to prediction, including any calculations.
  3. State the prediction as a testable claim about THIS problem's
     specific parameters.

CONSTRAINT: If a prediction cannot be traced to a stated principle,
you must EITHER add a new principle in Phase 1 or drop the prediction.
Predictions may not introduce new causal reasoning.

PHASE 3 -- CONCLUSION (final answer, synthesis only, no new reasoning)

Assemble your predictions into the answer format requested by the
question. The conclusion must:
  1. Follow NECESSARILY from the predictions -- no new logic.
  2. Include ALL fields and sections requested by the question.
  3. Cite which predictions support each part of the answer.

Wrap your response in this JSON structure, with the full answer inside
"conclusion" -> "answer":

{{
  "principles": [
    {{"id": "P1", "statement": "...",
      "type": "equilibrium|comparative-static|information-theoretic|
               strategic-interaction"}},
    {{"id": "P2", "statement": "...", "type": "..."}}
  ],
  "predictions": [
    {{
      "id": "PRED1",
      "statement": "...",
      "derived_from": ["P1"],
      "derivation": "step-by-step logical chain from principle to prediction"
    }}
  ],
  "conclusion": {{
    "answer": {{ COMPLETE answer with ALL requested fields }},
    "derived_from": ["PRED1", "PRED2"],
    "derivation": "how conclusion follows from predictions"
  }}
}}

=== QUESTION ===
{question}
\end{verbatim}
\end{small}

\subsection*{Adversarial Stress-Test Protocol}
After the model produces an initial response, this second-turn prompt is sent (\texttt{{response}} contains the initial answer):

\begin{small}
\begin{verbatim}
=== STRESS-TEST YOUR RESPONSE ===

You provided this response:

{response}

Now rigorously stress-test your reasoning:

1. ASSUMPTIONS -- list every assumption (explicit and implicit).
   Rate each: ROBUST / MODERATE / FRAGILE.
   Which assumption, if wrong, would most change your answer?

2. PARAMETER SENSITIVITY -- which values are load-bearing?
   At what threshold would your conclusion flip?

3. COUNTERFACTUAL CHECK -- what scenario would make your answer WRONG?
   Under what conditions would the OPPOSITE conclusion be correct?

4. CONFIDENCE -- rate HIGH / MEDIUM / LOW.
   What would increase or decrease your confidence?

5. REVISION DECISION -- based on the above analysis:
   - If you identified critical fragilities or LOW confidence, you SHOULD
     revise.
   - If revising, provide a COMPLETE new answer addressing the identified
     issues.

Respond in JSON:
{{
  "assumptions": [
    {{"assumption": "...", "fragility": "robust|moderate|fragile",
      "if_wrong": "..."}}
  ],
  "most_critical_assumption": "...",
  "parameter_sensitivity": [
    {{"parameter": "...", "flip_threshold": "..."}}
  ],
  "counterfactual": {{
    "what_would_make_wrong": "...",
    "conditions_for_opposite": "..."
  }},
  "confidence": {{
    "rating": "high|medium|low",
    "reasoning": "..."
  }},
  "revision_decision": {{
    "should_revise": true or false,
    "reason": "...",
    "revised_answer": null or {{ complete revised answer }},
    "revision_explanation": null or "what changed and why"
  }}
}}
\end{verbatim}
\end{small}

\subsection*{Contradiction Detection Protocol}
After all questions in a component are answered, the detection prompt is sent:

\begin{small}
\begin{verbatim}
=== CONTRADICTION DETECTION ===

You answered multiple questions in sequence. Now check your answers for
logical consistency.

YOUR ANSWERS:
{answers_text}

TASK:
1. Extract the KEY CLAIM from each answer (the main conclusion or principle).
2. For every pair of claims, assess:
   - CONSISTENT: both can be true simultaneously
   - TENSION: mild tension but not outright contradiction
   - CONTRADICTION: both cannot be true
3. List any contradictions found.

Respond in JSON:
{{
  "claims": [
    {{"question": "Q_ID", "claim": "the key claim"}}
  ],
  "pairwise_checks": [
    {{"q1": "Q_ID", "q2": "Q_ID",
      "status": "consistent|tension|contradiction",
      "explanation": "why"}}
  ],
  "contradictions_found": [
    {{"questions": ["Q_ID", "Q_ID"], "nature": "description"}}
  ],
  "has_contradictions": true or false
}}
\end{verbatim}
\end{small}

If contradictions are found, the revision prompt follows:

\begin{small}
\begin{verbatim}
=== CONTRADICTION REVISION ===

You identified these contradictions in your previous answers:

CONTRADICTIONS:
{contradictions_text}

ORIGINAL ANSWERS:
{original_answers_text}

TASK -- revise minimally to eliminate contradictions:
1. Make the SMALLEST changes necessary.
2. Keep correct answers intact where possible.
3. When two answers conflict, determine which is more defensible and
   revise the other.
4. Justify every revision.

Respond in JSON:
{{
  "analysis": "which answers to revise and why",
  "revisions": [
    {{
      "question_id": "Q_ID",
      "original_summary": "brief summary of original answer",
      "revised_answer": {{complete answer in original format}},
      "justification": "why this resolves the contradiction"
    }}
  ],
  "final_consistent_answers": {{
    "Q_ID": {{complete final answer}},
    "Q_ID": {{complete final answer}}
  }},
  "consistency_verification": "explanation of why the answers are now
  consistent"
}}
\end{verbatim}
\end{small}

\end{document}